\documentclass[10pt,conference,a4paper]{IEEEtran}
\usepackage{graphicx}
\usepackage{xcolor}
\usepackage{subcaption}
\usepackage{adjustbox}
\usepackage{multirow}
\usepackage{balance}
\usepackage{hyperref}

\newcommand{\hide}[1]{}

\newcommand{\rev}[1]{\textcolor{black}{#1}}
\newcommand{\comment}[1]{}
\ifCLASSINFOpdf
\else
\fi
\usepackage{amsmath}
\usepackage{amssymb}

\usepackage{bbold}
\usepackage{mathtools}

\usepackage{soul}

\usepackage{url}
\begin{document}

%%%%%%%%% TITLE
\title{Knowledge Distillation for Action Anticipation\\ via Label Smoothing}

%\author{\IEEEauthorblockN{Michael Shell}
%\IEEEauthorblockA{School of Electrical and\\Computer Engineering\\
%Georgia Institute of Technology\\
%Atlanta, Georgia 30332--0250\\
%Email: http://www.michaelshell.org/contact.html}
%\and
%\IEEEauthorblockN{Homer Simpson}
%\IEEEauthorblockA{Twentieth Century Fox\\
%\and
%\IEEEauthorblockN{James Kirk\\ and Montgomery Scott}
%\IEEEauthorblockA{Starfleet Academy\\
%San Francisco, California 96678--2391\\


\author{\IEEEauthorblockN{Guglielmo Camporese\IEEEauthorrefmark{1},
Pasquale Coscia\IEEEauthorrefmark{1},
Antonino Furnari\IEEEauthorrefmark{2},
Giovanni Maria Farinella\IEEEauthorrefmark{2},
Lamberto Ballan\IEEEauthorrefmark{1}}
\IEEEauthorblockA{\IEEEauthorrefmark{1}Department of Mathematics ``Tullio Levi-Civita'', University of Padova, Italy}
%Email: {\tt\small \{pasquale.coscia,lamberto.ballan\}@unipd.it}}
%Web: \url{http://vimp.math.unipd.it/}}
\IEEEauthorblockA{\IEEEauthorrefmark{2}Department of Mathematics and Computer Science, University of Catania, Italy}}


\maketitle

%%%%%%%%% ABSTRACT
\begin{abstract}
Human capability to anticipate near future from visual observations and non-verbal cues is essential for developing intelligent systems that need to interact with people. Several research areas, such as human-robot interaction (HRI), assisted living or autonomous driving need to foresee future events to avoid crashes or help people. Egocentric scenarios are classic examples where action anticipation is applied due to their numerous applications.
Such challenging task demands to capture and model domain's hidden structure to reduce prediction uncertainty. Since multiple actions may equally occur in the future, we treat action anticipation as a multi-label problem with missing labels extending the concept of label smoothing. This idea resembles the knowledge distillation process since useful information is injected into the model during training. We implement a multi-modal framework based on long short-term memory (LSTM) networks to summarize past observations and make predictions at different time steps. We perform extensive experiments on EPIC-Kitchens and EGTEA Gaze+ datasets including more than 2500 and 100 action classes, respectively. The experiments show that label smoothing systematically improves performance of state-of-the-art models for action anticipation.
\end{abstract}


% \fi
\IEEEpeerreviewmaketitle

%%%%%%%%% BODY TEXT
%------------------------------------------------------------------
\section{Introduction}
Human action analysis is a central task in computer vision that has a enormous impact on many applications, such as, video content analysis~\cite{ballan2011event,karpathy2014large}, video surveillance~\cite{3DCNN,Sultani_2018_CVPR}, and \rev{intelligent transportation}%automated driving vehicles
~\cite{Bhattacharyya_2018_CVPR, Maqueda_2018_CVPR}.
Systems interacting with humans also need the capability to promptly react to the context changes, and plan their actions accordingly.
Most previous works focus on the tasks of action recognition \cite{laptev2008learning,simonyan2014two,feichtenhofer2016convolutional} or early-action recognition \cite{ryoo2011human,cao2013recognize,de2016online}, i.e., recognition of an action \textit{after} its observation (happened in the past) or recognition of an \textit{on-going} action from its partial observation (only part of the current action is available). A more challenging task is to predict near future, i.e., to forecast actions that will be performed ahead in time.
Predicting future actions before observing the corresponding frames~\cite{Damen2020Collection,furnari2020rulstm} is demanded by many applications which need to anticipate human behaviour. For example, intelligent surveillance systems may support human operators to avoid hazards or assistive robotics may help non-self-sufficient people. Such task requires to analyze significant spatio-temporal variations among actions performed by different people. For this reason, multiple modalities (e.g., appearance and motion) are typically considered to improve the identification of similar actions.
Egocentric scenarios provide useful settings to study early-action recognition or action anticipation tasks. Indeed, wearable cameras offer an explicit point-of-view to capture human motion and object interaction.

\begin{figure}
    \centering
    \includegraphics[width=0.46\textwidth]{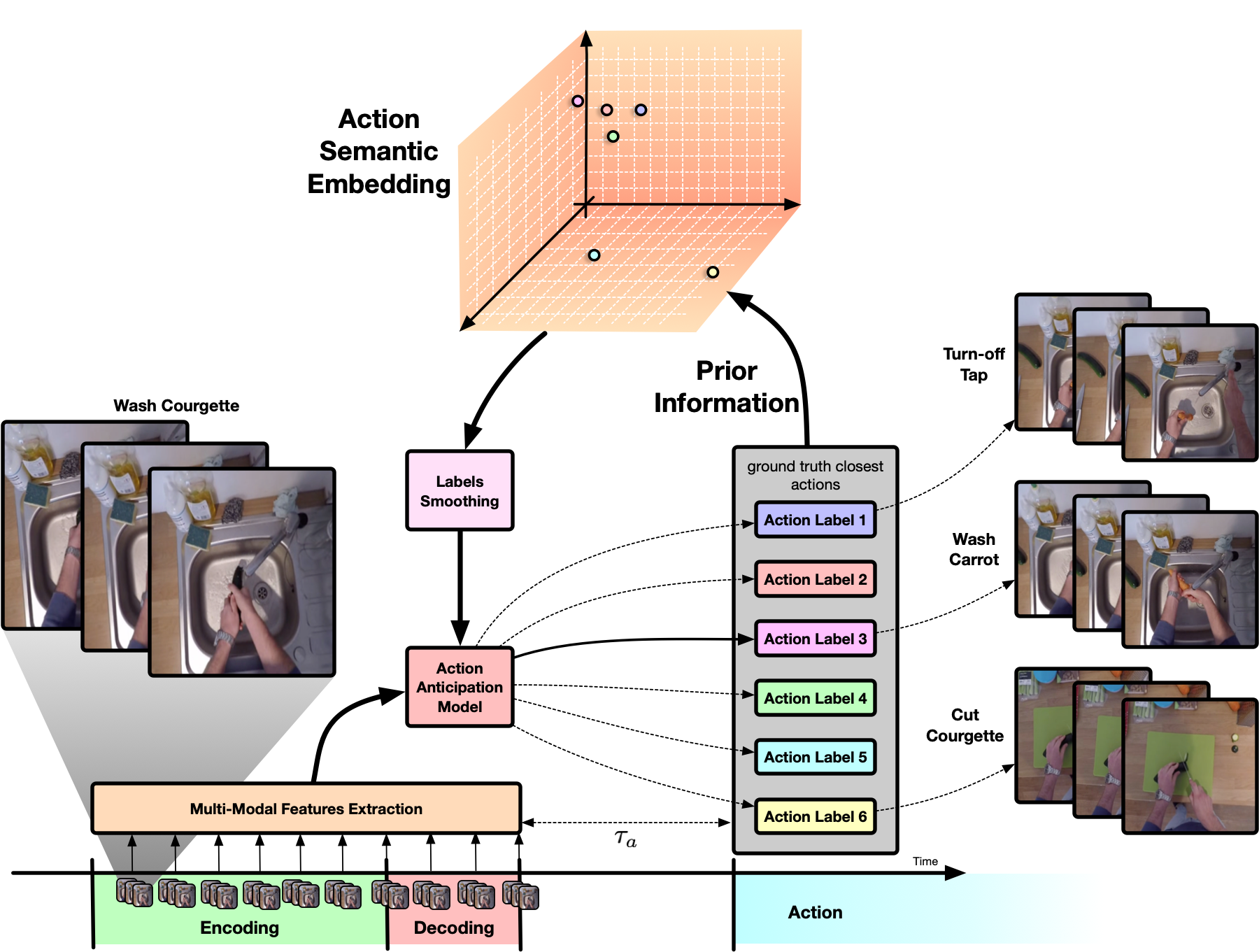}
    \caption{\rev{To anticipate next actions in egocentric scenarios, our framework processes input videos summarizing useful information during the encoding stage. In the decoding stage, our model predicts the next action at different timesteps.
    %Afterwards, it proposes the most plausible action that will occur in the future at different time-steps, in the decoding stage.
    Label smoothing techniques are employed to reduce the uncertainty on future predictions. }}%Our framework for action anticipation. After an encoding procedure of the video, during the decoding stage our model anticipates the next action that will occur in $\tau_{a}$ seconds. Afterwards, a teacher model distills semantic information via label smoothing into the action anticipation model during training in order to reduce the uncertainty on the future predictions.}
    \label{fig:intro}
\end{figure}

In this work, we address the problem of anticipating egocentric human actions in an indoor scenario at several time steps. More specifically, we \rev{aim to} anticipate an action by leveraging \rev{its previous} video segments. %that precede the action. We disentangle the processing of the video into encoding and decoding stages. 
\rev{We employ the encoding-decoding scheme of  \cite{furnari2020rulstm}. During the first stage, the model summarizes the video content extracting relevant information to infer the future while, in the second stage, predicts the next action at multiple 
time-steps (e.g., $1.5 s - 1.0 s - ... - 0.25 s$ before the action to be predicted).
Fig.~\ref{fig:intro} depicts such procedure. We exploit an LSTM-based network to capture temporal correlations among video frames and three different modalities as input representation: \textit{appearance} (RGB), \textit{motion} (optical flow) and \textit{object-based} features.}
%the first stage, the model summarizes the video content while in the second stage the model predicts at multiple anticipation times $\tau_a$ the next action (see Fig.~\ref{fig:intro}). We exploit a recurrent neural network (RNN) to capture temporal correlations between subsequent frames and consider three different modalities for representing the input: appearance (RGB), motion (optical flow) and object-based features.

%An important aspect to consider when dealing with human action anticipation is that the future is uncertain, which means that different prediction of future actions are equally likely to occur.
\rev{An important aspect to consider when dealing with human actions is the future's uncertainty since multiple actions may equally occur.}
For example, the actions ``\textit{sprinkle over pumpkin seeds}'' and ``\textit{sprinkle over sunflower seeds}''  may be equally performed when preparing a recipe. \rev{Nevertheless, assigning zero-probability to uncorrect yet semantically similar actions may lead predictions to not capture data structure randomly selecting from a set of plausible targets.}

For this reason, to deal with the uncertainty of future predictions, we propose \rev{several label smoothing techniques which}
%to group similar actions comparing \rev{their semantic meaning and use } several label smoothing techniques in order to 
broaden the set of possible futures and reduce the uncertainty caused by one-hot encoded labels. Label smoothing is introduced in~\cite{7780677} as a form of regularization for classification models since it introduces a positive constant value into wrong classes components of \rev{one-hot encoded targets.} %the one-hot target. 
%A peculiar feature of such method is to make models robust to overfitting especially when labels in the dataset are noisy, e.g., the targets are ambiguous. 
\rev{Compared to the typical knowledge distillation process where soft-labels are provided by an external network (the teacher) to train a smaller network (the student), our model 
distills knowledge extracted from prior of action labels (e.g., verb-noun). We extensively test our architecture on EPIC-Kitchens and EGTEA Gaze+ datasets which record egocentric scenarios where different people are involved to complete a recipe performing numerous actions and involving different objects within a kitchen. Our results show that label smoothing increases the performance of state-of-the-art models and yields better generalization on test data.}

%In our work, we extend label smoothing by using them as a bridge for distilling knowledge into the model during training. Our experiments on the large-scale EPIC-Kitchens dataset show that label smoothing increases the performance of state-of-the-art models and yields better generalization on test data.

The main contributions of our work are as follows: 1) we generalize the label smoothing idea extrapolating semantic priors from the action labels to capture the multi-modal future component of the action anticipation problem. We show that label smoothing, in this context, can be seen as a knowledge distillation process where the teacher gives semantic prior information for the action anticipation model; 2) we perform extensive experiments on egocentric videos proving that our label smoothing techniques systematically improve results of action anticipation of state-of-the-art models.

%------------------------------------------------------------------
\section{Related Work}
\label{sec:related}
Action anticipation requires the interpretation of the current activity using a number of observations in order to foresee the most likely actions. For this reason, we briefly review three related research areas: \textit{action recognition, early-action recognition} and \textit{action anticipation}.
\medskip

\textbf{\textit{Action recognition. }} 
Action recognition is the task of recognizing the action contained in an observed trimmed video. Classic approaches to action recognition have leveraged hand-designed features, coupled with machine learning algorithms to recognize actions from videos~\cite{ laptev2008learning,wang2013dense,wang2013action}.
More recent works have investigated the use of deep learning to obtain representations suitable for action recognition directly from videos in an end-to-end fashion.
Among these approaches, a line of works has investigated ways to exploit standard 2D CNNs for action recognition, often relying on optical flow as a mean to represent motion~\cite{simonyan2014two,feichtenhofer2017spatiotemporal,feichtenhofer2016convolutional,wang2016temporal,zhou2018temporal,lin2018temporal}.
Other works have focused on the extension of 2D CNNs to 3D CNNs able to process spatio-temporal volumes~\cite{karpathy2014large,tran2015learning,tran2018closer}.
Some approaches have used recurrent networks to model the temporal relationships between per-frame observations~\cite{donahue2015long,sudhakaran2018lsta,furnari2020rulstm}. All of these works investigate deeply how to represent and leverage the input video but less or even no importance is given to the representation of the action labels.
\medskip

\textbf{\textit{Early-action recognition.}}
Early action recognition consists in recognizing on-going actions from partial observations of streaming video~\cite{de2016online}.
Classic works have addressed the task using integral histograms of spatio-temporal features~\cite{ryoo2011human}, sparse-coding~\cite{cao2013recognize}, Structured Output SVMs~\cite{hoai2014max}, and Sequential Max-Margin event detectors~\cite{huang2014sequential}.
Another line of research has leveraged the use of LSTMs~\cite{aliakbarian2017encouraging,becattini2017done,de2018modeling,furnari2020rulstm} to account for the sequential nature of the task.
\medskip

\textbf{\textit{Action anticipation.}} Action anticipation deals with forecasting actions that will happen in the future. Previous studies have investigated different approaches such as hierarchical representations~\cite{lan2014hierarchical}, auto-regressive HMMs~\cite{jain2015car}, regressing future representations~\cite{vondrick2016anticipating}, encoder-decoder LSTMs~\cite{gao2017red}, and inverse reinforcement learning~\cite{zeng2017visual}.
Some other approaches proposed to perform long-term predictions only focusing on appearance features~\cite{Mahmud2017ICCV,Farha2018CVPR}. However, differently from this work, very little or even no attention has been payed either to the knowledge distillation of the action semantics or label smoothing for action anticipation.
\medskip

\textbf{\textit{Knowledge distillation and label smoothing}}
\comment{\gmf{do we have to discuss label smoothing works? is there any reference we can include? do we want to talk about the uncertancy of the future prediction, and hence our past work [1]?}}
Knowledge distillation \cite{44873} is the procedure of transferring the information extracted by a teacher network (with high learning capacity) to a student network (with low learning capacity) in order to allow the latter to reach similar performance. This is usually obtained by training the student via a distillation loss which takes into account both the ground truth and the prediction of the pre-trained teacher.
\rev{Since this procedure can distill useful information to low learning capacity networks, we perform semantic distillation via label smoothing without employing an external network that supervises the process.}

Label smoothing, introduced in \cite{7780677}, is the procedure of softening the distribution of the target labels, reducing the most confident value of the one-hot vector and considering a uniform value for all the zero vector components. Although such procedure improves results for classification problems reducing overfitting, no previous works investigate other design approaches except for the uniform smoothing. In our work, we both  generalize this idea and show a systematically improvement to state-of-the-art models. 

%\pasquale{Most works~\af{\cite{gao2017red,jain2016recurrent}} are based on long short-term memory (LSTM) architectures to capture temporal correlations (include refs.), or decouple video summarization and action prediction during training~\cite{furnari2020rulstm}. Some approaches propose long-term predictions focusing only on appearance features~\cite{Mahmud2017ICCV, Farha2018CVPR}. We extend such approaches introducing multi-modal features and redefining loss function to smooth (hard?) predictions.}

%%%%%%%%%%%%%%%%%%%%%%%%%%%%%%%%%%%%%%%%%%%%%
% Label Smoothing Papers References
%%%%%%%%%%%%%%%%%%%%%%%%%%%%%%%%%%%%%%%%%%%%%

% - REGULARIZING NEURAL NETWORKS BY PENALIZING CONFIDENT OUTPUT DISTRIBUTIONS ICLR 2017
% - When Does Label Smoothing Help? \cite{NIPS2019_8717}
% - Regularization via Structural Label Smoothing
% - Distilling the Knowledge in a Neural Network NIPS 2014 \cite{44873}
% - Rethinking the Inception Architecture for Computer Vision \cite{7780677}

%------------------------------------------------------------------
\section{Proposed Approach}
\label{sec:model}

Anticipating human actions is essential for developing intelligent systems able to avoid accidents or guide people to correctly perform their actions. \rev{Nevertheless, training models on one-hot encoded labels in egocentric videos may led to poor performance when dealing with similar actions. To this end, in the next sections we briefly describe our architecture and then study different label smoothing techniques to address this issue and improve generalization performance.}
%We study the suitability of label smoothing techniques to address the issue.
%To address this problem, we propose a multi-modal architecture based on RNNs.

%The proposed method captures temporal correlations between actions and label smoothing techniques to focus on a broader set of possible future actions.

%TOTO: add short intro to task protocol and figure:
%... our approach uses the protocol depicted in Fig.~\ref{fig:pipeline}

\begin{figure}[ht]
    \centering
    \begin{subfigure}[h]{0.25\textwidth}
        \centering
        \includegraphics[width=0.9\textwidth]{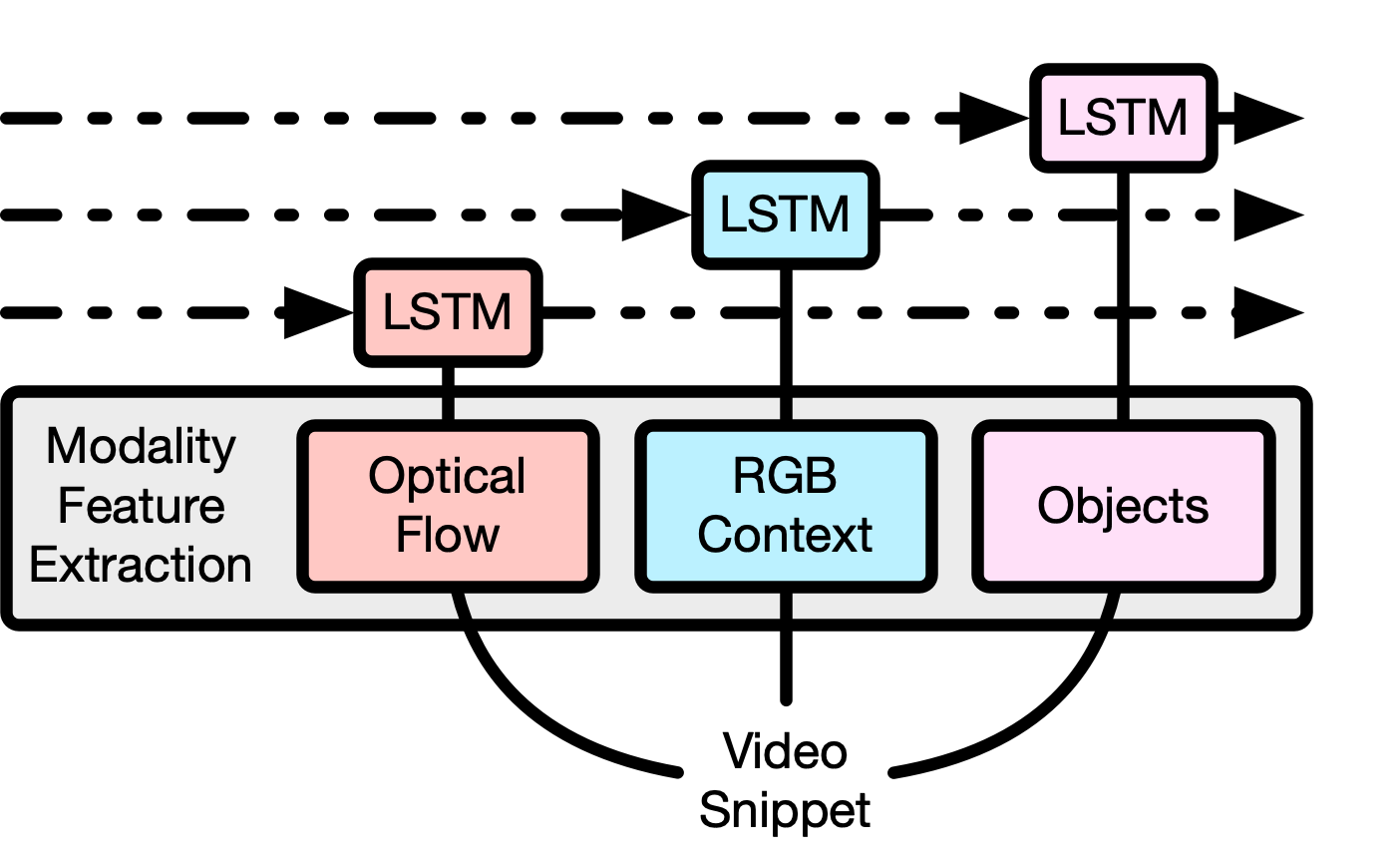}
        %\caption{Encoding block of the model.}
        \label{fig:model_block_E}
    \end{subfigure}%
    %~ 
    \begin{subfigure}[h]{0.25\textwidth}
        \centering
        \includegraphics[width=0.9\textwidth]{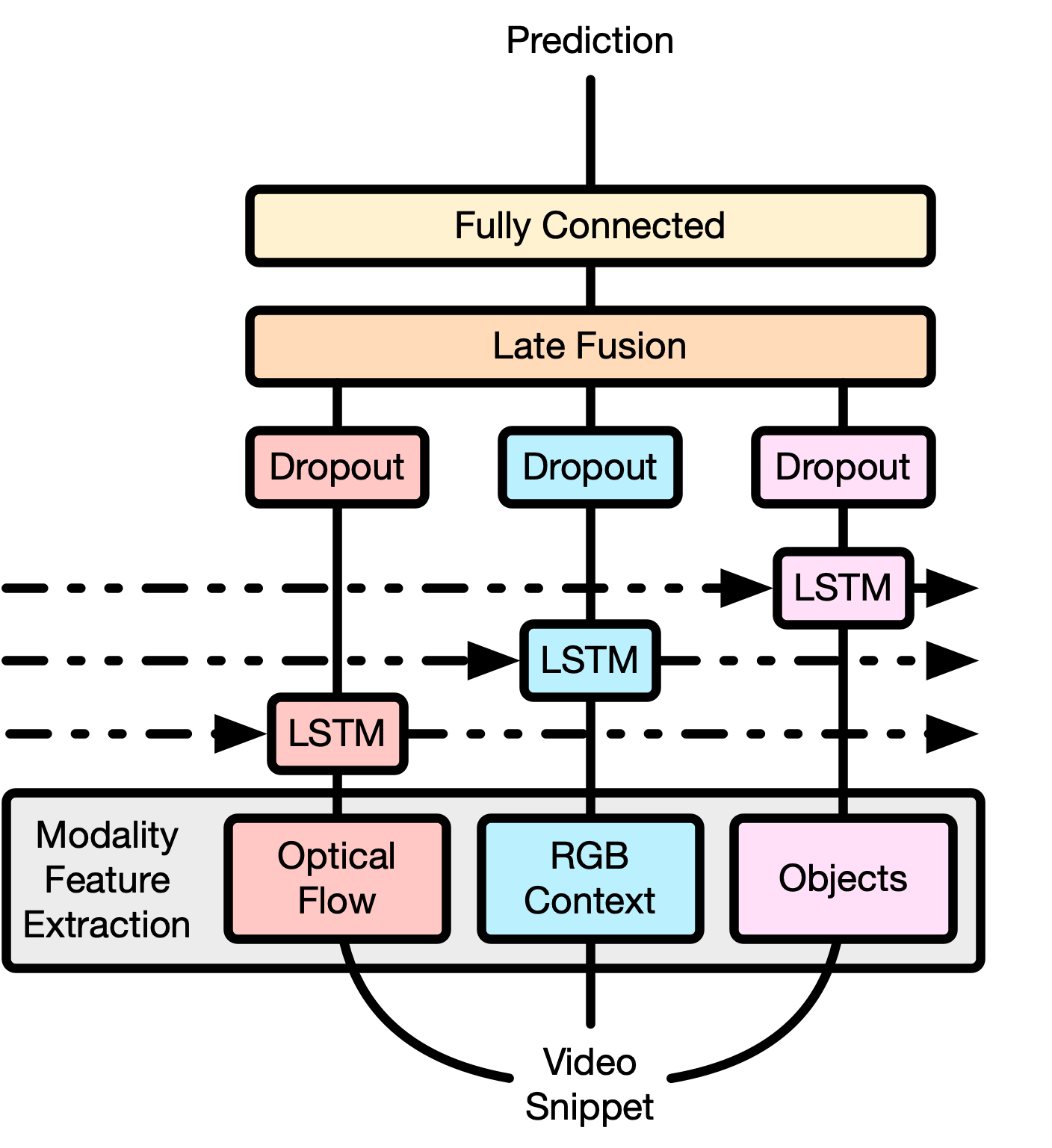}
        %\caption{Decoding block of the model.}
        \label{fig:model_block_D}
    \end{subfigure}
    \caption{\rev{Encoding (left) and decoding (right) branches of our architecture. The encoding branch processes, using LSTM networks,  video snippets of three modalities (optical flow, RGB and objects). The decoding branch simultaneously processes input video snippets and predicts the next action through a fully connected layer with soft-max activation.}%Encoding (left) and decoding (right) of our architecture. The encoding block takes as input the video snippet represented by three modalities (optical flow, RGB and objects). Each modality is separately fed into an LSTM. % of 1024 units that is unrolled in the encoding stage. 
    %The decoding block is composed by three LSTM branches % of units 1024 
    %that take as input the processed video snippets. We apply \rev{dropout} at each branch. % with rate 0.8 
    %Finally, all the branches are concatenated and passed to a fully connected layer with softmax activation.
    }
    \label{fig:model_architecture}
\end{figure}

%%%%%%%%%%%%%%%%%%%%%%%%%%%%%%%%%%%%%%%%%%%%%
% Implementation Details
%%%%%%%%%%%%%%%%%%%%%%%%%%%%%%%%%%%%%%%%%%%%%

\subsection{Action Anticipation Architecture}
\label{sec:model_architecture}
For our experiments, we consider a learning architecture based on recurrent neural networks (\rev{RNNs}).
%Following~\cite{furnari2020rulstm}, our approach uses the protocol depicted in Fig.~\ref{fig:pipeline} for anticipating future actions. 
We process \rev{input}
%the
frames preceding the action that we want to anticipate grouping them  into video snippet of length $5$. Each video snippet is collected every $\alpha=0.25$ seconds and processed considering three different modalities: \textit{RGB} features computed using a Batch Normalized Inception CNN \cite{pmlr-v37-ioffe15} trained for action recognition, \textit{objects} features computed using Fast-R CNN \cite{DBLP:journals/corr/Girshick15} and \textit{optical flow} computed as in \cite{wang2016temporal}, processed through a Batch Normalized Inception CNN trained for action recognition. Our multi-modal architecture processes the above inputs and encompasses two building blocks: an encoder which recognizes and summarises past observations and a decoder which predicts future actions at different anticipation time steps. \rev{Fig.~\ref{fig:model_architecture} depicts the two stages. Firstly, each modality is independently processed by an LSTM layer; then, such streams are merged with late fusion and fed to a fully connected layer to predict the next action.}%As shown in Fig.~\ref{fig:model_architecture}, during the encoding stage each modality is separately processed by a LSTM layer. During the decoding stage such streams are then merged with late fusion and fed into a fully connected layer using softmax activation. %Predictions are conditioned both on previous and current video snippets.

\begin{figure*}[ht]
    \centering
    \includegraphics[width=0.7\textwidth]{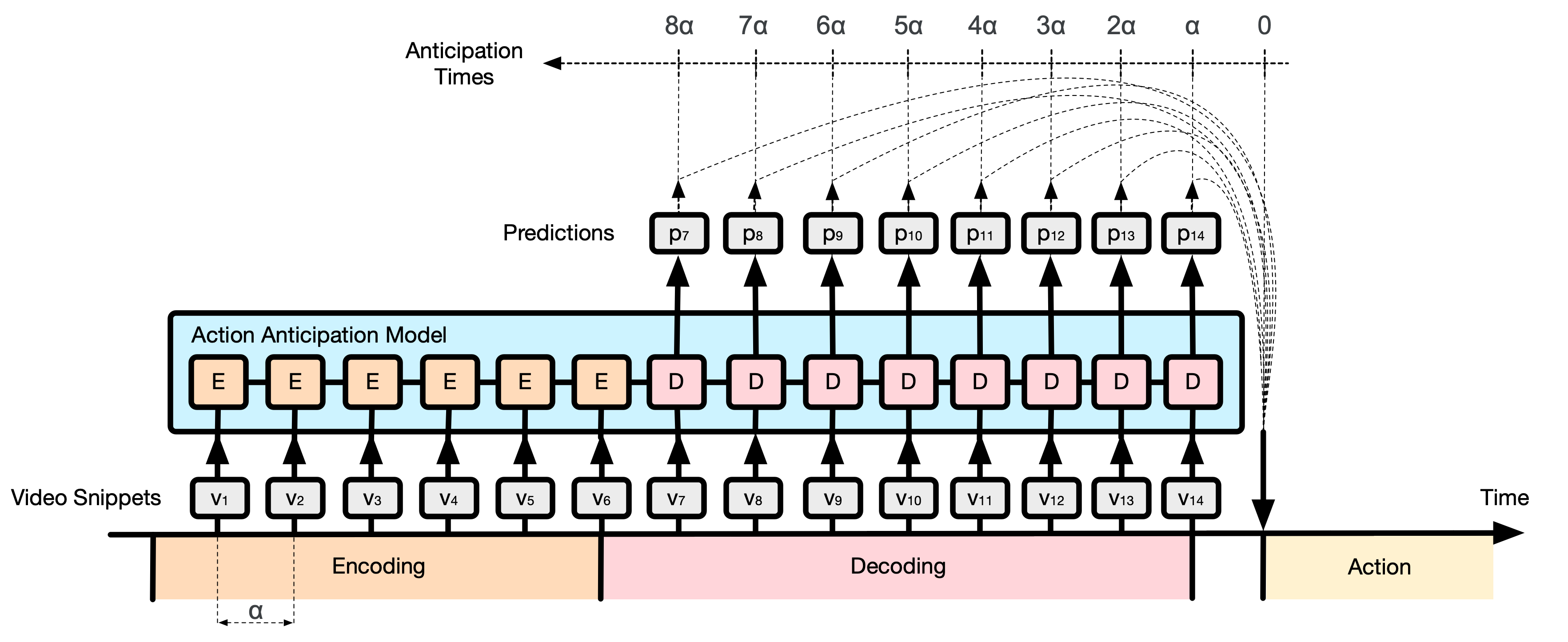}
    \caption{Action anticipation protocol based on the encoding and decoding stages. We summarize past observations by processing video snippets sampled every $\alpha = 0.25$ seconds in the encoding stage. After $6$ steps, we start making predictions every $\alpha$ seconds for $8$ steps to anticipate the future action. }
    %We encode temporal correlations between subsequent actions every $\alpha$ seconds from observed video snippets. After 6 time steps, we predict future action at $T_{act}$, every $\alpha$ seconds exploiting summarized data and current activity.}
    \label{fig:pipeline}
\end{figure*}

%%%%%%%%%%%%%%%%%%%%%%%%%%%%%%%%%%%%%%%%%%%%%
% Label Smoothing Design
%%%%%%%%%%%%%%%%%%%%%%%%%%%%%%%%%%%%%%%%%%%%% and to capture the future uncertainty

\subsection{Label Smoothing}
\label{sec:label_smoothing}
\rev{Egocentric videos exhibit an inherent uncertainty when dealing with predicting future actions~\cite{Furnari_2018_ECCV_Workshops}.}
%As investigated in \cite{Furnari_2018_ECCV_Workshops}, there is an inherent uncertainty on predicting future actions. 
In fact, \rev{given} %starting from 
the current state observation of an action there can be multiple, but still plausible, future scenarios that can occur. \rev{For this reason, the problem can be reformulated as a multi-label task with missing labels where, from a set of valid future realizations, only one is sampled.}

All previous models designed for action anticipation are trained with cross-entropy using one-hot labels, leveraging only one of the possible future scenarios as ground truth. A major drawback of using hard labels is to favour logits of correct classes weakening the importance of other plausible ones. In fact, given the one-hot encoded vector $y^{(k)}$ for a class $k$, the prediction of the model $p$ and the logits of the model $z$ such that $p(i) = e^{z(i)} / \sum_j e^{z(j)}$, the cross entropy is minimized only if $z(k) \gg z(i) \ \forall i \neq k$. This fact encourages the model to be over-confident about its predictions since during training it tries to focus all the energy on one single logit leading to overfitting and scarce adaptivity \cite{7780677}. \rev{To this purpose, we smooth the target distribution enabling the chance of negative (yet still plausible) classes to be selected}. \rev{The most common form of label smoothing procedure relies on a uniform positive component added to a set of similar classes without capturing the difference between such actions. To overcome this issue, we  design and compare several label smoothing techniques that exploit both semantic and temporal relations among actions. In the following, we firstly model the general idea of label smoothing and then propose multiple techniques to define the corresponding components.}%However, the usual label smoothing procedure introduces a uniform positive component among all the classes, without capturing the difference between actions. To this end, we propose several ways of designing such smoothing procedure by encoding semantic priors into the labels and weighting the actions according their feature representation.
%We can connect our soft labels approach to the knowledge distillation framework~\cite{44873} where the teacher provides useful information to the student model during training. What differs is that the teacher does not depend on the input data but solely on the target, i.e., it distills information using ground-truth data. Since teacher's prediction is constant w.r.t. the input, such information can be encoded before training into the target via label smoothing.

As a form of regularization, \cite{7780677} introduces the idea of smoothing hard labels by averaging one-hot encoded targets with constant vectors as follows: 
\begin{equation}
  y^{soft}(i) = (1-\alpha) y(i)  + \alpha/K,
\label{eq_softlabel}
\end{equation}
where \rev{$y(i)$ is the one-hot encoding of the $i^{th}$ example}, $\alpha$ is the smoothing factor ($0\leq \alpha \leq 1$) and $K$ represents the number of classes.
Since cross entropy is linear w.r.t. its first argument, it can be written as follows:
\begin{equation}
    \begin{split}
        CE[y^{soft}, p] = \sum_i - y^{soft}(i)\log(p(i)) = \\
        = (1 - \alpha) CE[y, p] + \alpha CE[1/K, p].
    \end{split}
    \label{eq:xent}
\end{equation}
The optimization based on the above loss can be seen as a distillation knowledge procedure \cite{44873} where the teacher randomly predicts the output, i.e., $p^{teacher}(i) = 1/K, \ \forall i$. Hence, the connection with the distillation loss proves that the second term in Eq.~(\ref{eq_softlabel}) can be seen as a prior knowledge, given by an agnostic teacher, for the target $y$. %Although using an agnostic teacher seems an unusual choice, uniform label smoothing can be seen as a form of regularization \cite{7780677} and thus it can improve the model's generalization capability. 
Taking this into account, we extend the idea of smoothing labels by modeling the second term of Eq.~(\ref{eq_softlabel}), i.e., the prior knowledge of the targets, as follows:

\begin{equation}
    y^{soft}(i) = (1-\alpha) y(i)  + \alpha \pi(i),
    \label{eq:smooth_label_gen}
\end{equation}
where $\pi \in \mathbb{R}^K$ is the prior vector such that $\sum_i \pi(i) = 1$ and $\pi(i) \geq 0 \ \forall i$.

Therefore, the resulting cross entropy with soft labels is written as follows:

\begin{equation}
    CE[y^{soft}, p] = (1 - \alpha) CE[y, p] + \alpha CE[\pi, p]
    \label{eq:xent_prior}
\end{equation}

This loss not only penalizes errors related to the correct class but also errors related to the positive entries of the prior. Starting from this formulation, we introduce \textit{Verb-Noun}, \textit{GloVe} and \textit{Temporal} priors for smoothing labels in the knowledge distillation procedure. %In the following, we detail our label smoothing techniques.
\medskip

%%%%%%%%%%%%%%%%%%%%%%%%%%%%%%%%%%%%%%%%%%%%%
% Verb-Noun Label Smoothing
%%%%%%%%%%%%%%%%%%%%%%%%%%%%%%%%%%%%%%%%%%%%%

\textbf{\textit{Verb-Noun label smoothing.}}
EPIC-KITCHENS~\cite{Damen2020Collection} contains action labels structured as verbs-noun pairs, like ``\textit{cut onion}'' or ``\textit{dry spoon}''. More formally, if we define $\mathcal{A}$ the set of actions, $\mathcal{V}$ the set of verbs, and $\mathcal{N}$ the set of nouns, then an action is represented by a tuple $a = (v, n)$ where $v \in \mathcal{V}$ and $n \in \mathcal{N}$. Let $\mathcal{A}_v(\bar{v})$ the set of actions sharing the same verb $\bar{v}$ and $\mathcal{A}_n(\bar{n})$
the set of actions sharing the same noun $\bar{n}$, defined as follows:

\begin{equation}
    \mathcal{A}_v(\bar{v}) = \{(\bar{v}, n) \in \mathcal{A}\ \forall\ n \in \mathcal{N}\},
\end{equation}
\begin{equation}
    \mathcal{A}_n(\bar{n}) = \{(v, \bar{n}) \in \mathcal{A} \ \forall\ v \in \mathcal{V}\},
\end{equation}
where $\bar{n} \in \mathcal{N}$ and $\bar{v} \in \mathcal{V}$.

We define the prior of the $k^{th}$ ground-truth action class as

\begin{equation}
    %\begin{split}
        \pi_{VN}^{(k)}(i) =\mathbb{1}\left[  a^{(i)} \in \mathcal{A}_v(v^{(k)}) \cup \mathcal{A}_n(n^{(k)}) \right] \frac{1}{C_k},
    %\end{split}
\end{equation}
where $a^{(i)}$ is the $i^{th}$ action, $v^{(k)}$ and $n^{(k)}$ are the verb and the noun of the $k^{th}$ action, $\mathbb{1}[\cdot]$ is the indicator function, and $C_k = |\mathcal{A}_v(v^{(k)})| + |\mathcal{A}_n(n^{(k)})| - 1$ is a normalization term.
Using such encoding rule, the cross entropy not only penalizes the error related to the correct class but also the errors with respect to all the other \textit{similar} actions with either the same verb or noun\footnote{It can be proved that in terms of scalar product two different classes having the same noun or verb and encoded with Verb-Noun label smoothing are closer with respect to classes encoded with hard labels}. 
\medskip

%%%%%%%%%%%%%%%%%%%%%%%%%%%%%%%%%%%%%%%%%%%%%
% GloVe Based Label Smoothing
%%%%%%%%%%%%%%%%%%%%%%%%%%%%%%%%%%%%%%%%%%%%%
\textbf{\textit{GloVe based label smoothing.}}
An important aspect to consider when dealing with actions represented by verbs and/or nouns is their semantic meaning. In the \textit{Verb-Noun} label smoothing, we define the prior considering a rough yet still meaningful semantic where actions that share either the same verb or noun are considered similar. To extend this idea, we extrapolate the prior from the word embedding of the actions. One of the most important properties of word embeddings is to put closer words with similar semantic meanings and to move dissimilar ones far away, as opposed to hard labels that cannot capture at all similarity between classes %since $y^{(i)T}y^{(j)} = 0 \ \forall i\neq j$.

Using such action representation, we enable the distillation of useful information into the model during training since the cross entropy not only penalizes the error related to the correct class but also the error related to all other similar actions.
In order to compute the word embeddings of the actions we use the GloVe model \cite{pennington2014glove} pretrained on the Wikipedia 2014 and Gigaword 5 datasets. \rev{We use the GloVe model since it does not only rely on local statistics of words, but also incorporates global statistics to obtain word vectors.} Since the model takes as input only single words, we encode the action as follows: 

\begin{equation}
    \phi^{(k)} = Concat\left[ GloVe(v^{(k)}), GloVe(n^{(k)}) \right]
\end{equation}
\noindent
where $\phi$ is the obtained action representation of $a^{(k)} = (v^{(k)}, n^{(k)})$ and $GloVe(\cdot) \in \mathbb{R}^{300}$ is the output of the GloVe model. We finally compute the prior probability for smoothing the labels as the similarity between two action representations, which is computed as follows:

\begin{equation}
    \pi^{(k)}_{GL}(i) = \frac{|\phi^{(k)T} \phi^{(i)}|}{\sum_j |\phi^{(k)T} \phi^{(j)}|}.
    \label{eq:glove_sim}
\end{equation}

Hence, $\pi^{(k)}_{GL}(i)$ in Eq.~(\ref{eq:glove_sim}) represents the similarity between the $k^{th}$ and the $i^{th}$ action.
\medskip
%%%%%%%%%%%%%%%%%%%%%%%%%%%%%%%%%%%%%%%%%%%%%
% Temporal Label Smoothing
%%%%%%%%%%%%%%%%%%%%%%%%%%%%%%%%%%%%%%%%%%%%%

\textbf{\textit{Temporal label smoothing.}} %\st{`Since action anticipation relies on the prediction of future activity starting from past and current observations, we designed a prior starting from the sequence of transitions between consecutive actions. Intuitively if in the dataset a particular sequence of actions occur frequently then it is easier for the model to recognize it if we inject the pattern information into the training process.}
%Action anticipation relies on the prediction of future activity from past and current observations. 
%Some actions are more likely to co-occur than others. Furthermore, only specific action sequences may be considered plausible.
%For this reason, it could be reasonable to focus on most frequent action sequences since they may reveal possible valid paths in the actions space.
%In this case, we build the prior probability of their observation by considering subsequent actions of length two, i.e., we estimate from the training set the transition probability from the $i^{th}$ to the $k^{th}$ action as follows:
%\begin{equation}
%    \pi^{(k)}_{TE}(i) = \frac{ Occ\left[a^{(i)} \rightarrow a^{(k)} \right] }{\sum_j Occ\left[ a^{(j)} \rightarrow a^{(k)} \right]}
%\end{equation}
%where $Occ\left[a^{(i)} \rightarrow a^{(k)} \right]$ is the number of times that the $i^{th}$ action is followed by the $k^{th}$ action. Using such representation, we reward both the correct class and most frequent actions that precede the correct class.
\rev{Some pairs of actions are more likely to occur than other ones. Furthermore, only specific action sequences may be considered valid.}
For this reason, it could be reasonable to focus on most frequent action sequences since they may reveal possible valid paths in the actions space.
In this case, we build the prior probability of their observations by considering \rev{two subsequent actions in order to } estimate % from the training set
the transition probability from the $i^{th}$ to the $k^{th}$ action as follows:

\begin{equation}
    \pi^{(k)}_{TE}(i) = \frac{ Occ\left[a^{(i)} \rightarrow a^{(k)} \right] }{\sum_j Occ\left[ a^{(j)} \rightarrow a^{(k)} \right]}
\end{equation}

where $Occ\left[a^{(i)} \rightarrow a^{(k)} \right]$ is the number of times that the $i^{th}$ action is followed by the $k^{th}$ action. Using such representation, we reward both the correct class and most frequent actions that precede the correct class.
%%%%%%%%%%%%%%%%%%%%%%%%%%%%%%%%%%%%%%%%
% Table of the smooth factor
%%%%%%%%%%%%%%%%%%%%%%%%%%%%%%%%%%%%%%%%

\begin{table}[!t]
    \centering
    \begin{adjustbox}{width=\linewidth, center}
	\setlength{\tabcolsep}{3pt}
    \begin{tabular}{ c c c c c c}
        \hline
        %\multicolumn{6}{ c }{& Temporal & Uniform & Verb-Noun & GloVe & GLoVe+Verb-Noun}\\
        & Temporal & Uniform & Verb-Noun & GloVe & GLoVe+Verb-Noun \\
        \hline
        $\alpha^*$ & 0.6 & 0.1 & 0.45 & 0.6 & 0.5 \\
        \hline
    \end{tabular}
    \end{adjustbox}
    \caption{Results of a grid search procedure to detect the best smooth factor $\alpha$ for proposed label smoothing techniques.}
    \label{tab:alpha}
\end{table}

%%%%%%%%%%%%%%%%%%%%%%%%%%%%%%%%%%%%%%%%%%%%%
% Experiments
%%%%%%%%%%%%%%%%%%%%%%%%%%%%%%%%%%%%%%%%%%%%%
\section{Experiments}
\label{sec:experiments}
\subsection{\rev{Dataset, Experimental Protocol and Evaluation Measures}}
\textbf{\textit{Dataset.}}
\rev{Our experiments are performed on EPIC-KITCHENS~\cite{Damen2020Collection} and EGTEA Gaze+~\cite{Li2018InTE} datasets. The former,} is a large-scale collection of egocentric videos that contains $39,596$ action annotations divided into $2513$ unique actions, $125$ verbs, and $352$ nouns. We use the same split as \cite{furnari2020rulstm} producing $23,493$ segments for training and $4,979$ segments for validation. \rev{The latter contains $10,325$ action annotations, $19$ verbs, $5$1 nouns and $106$ unique actions. In EGTEA Gaze+, methods are evaluated reporting the average performance across the three splits provided by the authors of the dataset~\cite{Li2018InTE}.}
%Following \cite{furnari2020rulstm}, we split the public training set into training ($23,493$ segments) and validation ($4,979$ segments) sets by randomly choosing $232$ videos for training and $40$ videos for validation.
\medskip

\rev{\textbf{\textit{Experimental Protocol.}}
Following~\cite{furnari2020rulstm}, our approach uses the protocol depicted in Fig.~\ref{fig:pipeline} for anticipating future actions based on encoding and decoding stages. Given the action to predict, we consider the previous $14$ timesteps spaced out by $0.25~s$. In the encoding phase, our network processes input video snippets of $5$ frames to extract relevant information about the past ($6$ timesteps). During the decoding phase, it predicts the next action at $8$ successive timesteps simultaneously processing the corresponding video snippets. This protocol lets us to measure the performance of the network at different anticipation times.}
\medskip

\textbf{\textit{Evaluation Measures.}}
To asses the quality of predictions and compare all methods, we use the Top-k accuracy, i.e. we assume the prediction correct if the label falls into the best top-k predictions.
As reported in~\cite{Furnari_2018_ECCV_Workshops, koppula2016anticipating}, such measure is one of the most appropriate given the uncertainty of future predictions. More specifically, we use the Top-5 accuracy for methods comparison. For the test set, we also use the Top-1 accuracy, the Macro Average Class Precision and the Macro Average Class Recall. The last two metrics are computed only on many-shot nouns, verbs and actions as explained in~\cite{Damen2020Collection}.

%%%%%%%%%%%%%%%%%%%%%%%%%%%%%%%%%%%%%%%%
% Table our baseline
%%%%%%%%%%%%%%%%%%%%%%%%%%%%%%%%%%%%%%%%

\begin{table*}[!t]
    \centering
    \begin{adjustbox}{width=\linewidth,center}
	\setlength{\tabcolsep}{3pt}
    \begin{tabular}{ l c c c c c c c c} 
        
        \hline
        \multicolumn{9}{ c }{Top-5 Action Accuracy \% @ different anticipation times [$s$]}\\
        \hline
        & 2 & 1.75 & 1.5 & 1.25 & 1 & 0.75 & 0.5 & 0.25 \\
        \hline
        LSTM One-hot Encoding & $27.71 \pm \scriptstyle{0.33}$ & $28.69 \pm \scriptstyle{0.34}$ & $29.84 \pm \scriptstyle{0.24}$ & $30.90 \pm \scriptstyle{0.48}$ & $31.93 \pm \scriptstyle{0.45}$ & $33.14 \pm \scriptstyle{0.36}$ & $34.10 \pm \scriptstyle{0.44}$ & $35.16 \pm \scriptstyle{0.35}$ \\
        LSTM TE Smoothing & $27.94 \pm \scriptstyle{0.24}$ & $28.90 \pm \scriptstyle{0.27}$ & $30.06 \pm \scriptstyle{0.24}$ & $31.13 \pm \scriptstyle{0.19}$ & $32.19 \pm \scriptstyle{0.28}$ & $33.21 \pm \scriptstyle{0.36}$ & $34.17 \pm \scriptstyle{0.37}$ & $35.10 \pm \scriptstyle{0.25}$  \\
        LSTM Uniform Smoothing & $28.16 \pm \scriptstyle{0.27}$ & $29.06 \pm \scriptstyle{0.26}$ & $30.23 \pm \scriptstyle{0.24}$ & $31.25 \pm \scriptstyle{0.27}$ & $32.41 \pm \scriptstyle{0.28}$ & $33.64 \pm \scriptstyle{0.27}$ & $34.69 \pm \scriptstyle{0.19}$ & $35.75 \pm \scriptstyle{0.13}$ \\
        LSTM VN Smoothing & $28.43 \pm \scriptstyle{0.30}$ & $29.41 \pm \scriptstyle{0.31}$ & $30.68 \pm \scriptstyle{0.28}$ & $31.85 \pm \scriptstyle{0.21}$ & $33.08 \pm \scriptstyle{0.18}$ & $34.35 \pm \scriptstyle{0.19}$ & $35.38 \pm \scriptstyle{0.34}$ & $36.46 \pm \scriptstyle{0.26}$ \\
        LSTM GL Smoothing & $28.61 \pm \scriptstyle{0.26}$ & $29.87 \pm \scriptstyle{0.25}$ & $30.97 \pm \scriptstyle{0.34}$ & $31.94 \pm \scriptstyle{0.34}$ & $33.12 \pm \scriptstyle{0.36}$ & $34.40 \pm \scriptstyle{0.37}$ & $35.51 \pm \scriptstyle{0.37}$ & $36.87 \pm \scriptstyle{0.25}$ \\
        \textbf{LSTM GL+VN Smoothing} & $\mathbf{28.88} \pm \scriptstyle{0.20}$ & $\mathbf{29.94} \pm \scriptstyle{0.19}$ & $\mathbf{31.23} \pm \scriptstyle{0.32}$ & $\mathbf{32.54} \pm \scriptstyle{0.31}$ & $\mathbf{33.56} \pm \scriptstyle{0.28}$ & $\mathbf{34.92} \pm \scriptstyle{0.25}$ & $\mathbf{36.06} \pm \scriptstyle{0.33}$ & $\mathbf{37.29} \pm \scriptstyle{0.30}$ \\
        \hline
        Improv. & +1.17 & +1.25 & +1.39 & +1.64 & +1.63 & +1.78 & +1.96 & +2.13 \\
        \hline
        \hline
        RU-LSTM & $29.44$ & $30.73$ & $32.24$ & $33.41$ & $35.32$ & $36.34$ & $37.37$ & $38.39$ \\
        \textbf{RU-LSTM GL+VN Smoothing} & $\mathbf{30.37}$ & $\mathbf{31.64}$ & $\mathbf{33.17}$ & $\mathbf{34.86}$ & $\mathbf{35.90}$ & $\mathbf{37.07}$ & $\mathbf{38.96}$ & $\mathbf{39.74}$ \\
        \hline
        Improv. & $+0.93$ & $+0.91$ & $+0.93$ & $+1.45$ & $+0.58$ & $+0.73$ & $+1.59$ & $+1.35$ \\
        \hline
    \end{tabular}
    \end{adjustbox}
    \caption{Top-5 accuracy for action anticipation task at different anticipation time steps on EPIC-KITCHENS validation set. 
    We report results of several label smoothing techniques and show a systematic performance improvement compared to one-hot encoding. These results are confirmed for a simple LSTM-based model and also for the state-of-the-art RU-LSTM~\cite{furnari2020rulstm} model.}
    \label{tab:baseline}
\end{table*}

%%%%%%%%%%%%%%%%%%%%%%%%%%%%%%%%%%%%%%%%
% Table RULSTM Validation
%%%%%%%%%%%%%%%%%%%%%%%%%%%%%%%%%%%%%%%%

%\begin{table*}[!t]
%    \centering
%    \begin{adjustbox}{width=\linewidth, center}
%	\setlength{\tabcolsep}{11pt}
%    \begin{tabular}{ l c c c c c c c c} 
%       
%        \hline
%        \multicolumn{9}{ c }{Top-5 Action Accuracy \% @ different anticipation times [$s$]}\\
%        \hline
%       & 2 & 1.75 & 1.5 & 1.25 & 1 & 0.75 & 0.5 & 0.25 \\
%        \hline
%        RU-LSTM & $29.44$ & $30.73$ & $32.24$ & $33.41$ & $35.32$ & $36.34$ & $37.37$ & $38.39$ \\
%        \textbf{RU-LSTM Smooth GL + VN} & $\mathbf{30.37}$ & $\mathbf{31.64}$ & $\mathbf{33.17}$ & $\mathbf{34.86}$ & $\mathbf{35.90}$ & $\mathbf{37.07}$ & $\mathbf{38.96}$ & $\mathbf{39.74}$ \\
%        \hline
%        Improv. & $+0.93$ & $+0.91$ & $+0.93$ & $+1.45$ & $+0.58$ & $+0.73$ & $+1.59$ & $+1.35$ \\
%        \hline
%    \end{tabular}
%    \end{adjustbox}
%    \caption{Top-5 accuracy for action anticipation task at different anticipation time step on the EPIC-KITCHENS validation set. We introduce \textit{GL+VN} smoothing technique into state-of-the-art RU-LSTM~\cite{furnari2020rulstm} model.}
%    \label{tab:rulstm_val}
%\end{table*}

%%%%%%%%%%%%%%%%%%%%%%%%%%%%%%%%%%%%%%%%
% Table branches (EPIC-KITCHENS)
%%%%%%%%%%%%%%%%%%%%%%%%%%%%%%%%%%%%%%%%

\begin{table*}[!t]
    \centering
    \begin{adjustbox}{width=\linewidth,center}
    \setlength{\tabcolsep}{11pt}
    \begin{tabular}{ l c c c c c c c c} 
        
        \hline
        \multicolumn{6}{ c }{\rev{Top-5 Action Accuracy \% @ 1 [$s$]}}\\
        \hline
        & \rev{One-hot} & \rev{TE Smoothing} & \rev{Uniform Smoothing} & \rev{VN Smoothing} & \rev{GL Smoothing} \\
        \hline
        \rev{LSTM (RGB-only)} & $\rev{29.54}$  & $\rev{29.60}$  &  $\rev{29.76}$ &  $\rev{29.83}$ & $\mathbf{\rev{29.85}}$ \\
        \rev{LSTM (Flow-only)} & $\rev{20.43}$ &  $\rev{20.53}$ &  $\rev{20.65}$ &  $\rev{20.69}$ &  $\mathbf{\rev{20.77}}$ \\
        \rev{LSTM (OBJ-only)} & $\rev{29.5}$  &  $\rev{29.64}$ &  $\rev{29.69}$ &  $\rev{29.79}$ &  $\mathbf{\rev{29.82}}$ \\
        \hline
        \rev{RU-LSTM (RGB-only)} & $\rev{30.83}$  & $\rev{30.95}$ &  $\rev{31.00}$ & $\rev{31.05}$  & $\mathbf{\rev{31.19}}$ \\
        \rev{RU-LSTM (Flow-only)} & $\rev{21.42}$  & $\rev{21.51}$ &  $\rev{21.51}$ &  $\rev{21.63}$  & $\mathbf{\rev{21.73}}$ \\
        \rev{RU-LSTM (OBJ-only)} & $\rev{29.89}$  & $\rev{29.99}$ &  $\rev{30.07}$ &  $\rev{30.04}$  & $\mathbf{\rev{30.19}}$ \\

        \hline
    \end{tabular}
    \end{adjustbox}
    \caption{\rev{Top-5 accuracy for action anticipation task at $\tau_a=1$ second on the EPIK-KITCHENS validation set for each model branch (i.e. RGB, Flow and OBJ). 
    We report results of several label smoothing techniques and show a systematic performance improvement compared to one-hot encoded labels.}}
    \label{tab:branches}
\end{table*}

%%%%%%%%%%%%%%%%%%%%%%%%%%%%%%%%%%%%%%%%
% Table our baseline (EGTEA GAZE+)
%%%%%%%%%%%%%%%%%%%%%%%%%%%%%%%%%%%%%%%%

\begin{table*}[!t]
    \centering
    \begin{adjustbox}{width=\linewidth,center}
	\setlength{\tabcolsep}{11pt}
    \begin{tabular}{ l c c c c c c c c} 
        
        \hline
        \multicolumn{9}{ c }{\rev{Top-5 Action Accuracy \% @ different anticipation times [$s$]}}\\
        \hline
        & \rev{2} & \rev{1.75} & \rev{1.5} & \rev{1.25} & \rev{1} & \rev{0.75} & \rev{0.5} & \rev{0.25} \\
        \hline
        \rev{LSTM One-hot Encoding} & $\rev{55.94}$ & $\rev{58.75}$ & $\rev{60.94}$ & $\rev{63.02}$ & $\rev{65.78}$ & $\rev{68.04}$ & $\rev{71.55}$ & $\rev{73.94}$ \\
        \rev{LSTM TE Smoothing} & $\rev{56.13}$ & $\rev{58.93}$ & $\rev{61.17}$ & $\rev{63.24}$ & $\rev{66.00}$ & $\rev{68.20}$ & $\rev{71.75}$ & $\rev{74.20}$ \\
        \rev{LSTM Uniform Smoothing} & $\rev{56.35}$ & $\rev{59.20}$ & $\rev{61.37}$ & $\rev{63.36}$ & $\rev{66.12}$ & $\rev{68.41}$ & $\rev{71.95}$ & $\rev{74.35}$ \\
        \rev{LSTM VN Smoothing} & $\rev{56.85}$ & $\rev{59.67}$ & $\rev{61.64}$ & $\rev{64.03}$ & $\rev{66.81}$ & $\rev{68.94}$ & $\rev{72.56}$ & $\rev{75.44}$ \\
        \rev{\textbf{LSTM GL Smoothing}} & $\mathbf{\rev{57.34}}$ & $\mathbf{\rev{60.11}}$ & $\mathbf{\rev{62.25}}$ & $\mathbf{\rev{64.42}}$ & $\mathbf{\rev{67.21}}$ & $\mathbf{\rev{69.56}}$ & $\mathbf{\rev{73.02}}$ & $\mathbf{\rev{75.83}}$ \\
        \hline
        \rev{Improv.} & \rev{+1.4} & \rev{+1.36} & \rev{+1.31} & \rev{+1.4} & \rev{+1.43} & \rev{+1.52} & \rev{+1.47} & \rev{+1.89} \\
        \hline
        \hline
        \rev{RU-LSTM} & $\rev{56.82}$ & $\rev{59.13}$ & $\rev{61.42}$ & $\rev{63.53}$ & $\rev{66.40}$ & $\rev{68.41}$ & $\rev{71.84}$ & $\rev{74.28}$ \\
        \rev{\textbf{RU-LSTM GL Smoothing}} & $\mathbf{\rev{59.99}}$ & $\mathbf{\rev{62.02}}$ & $\mathbf{\rev{63.95}}$ & $\mathbf{\rev{66.47}}$ & $\mathbf{\rev{68.74}}$ & $\mathbf{\rev{72.16}}$ & $\mathbf{\rev{75.21}}$ & $\mathbf{\rev{78.11}}$ \\
        \hline
        \rev{Improv.} & \rev{+3.17} & \rev{+2.89} & \rev{+2.53} & \rev{+2.94} & \rev{+2.34} & \rev{+3.75} & \rev{+3.37} & \rev{+3.83} \\
        \hline
    \end{tabular}
    \end{adjustbox}
    \caption{\rev{Top-5 accuracy for action anticipation task at different anticipation time steps on the EGTEA GAZE+ dataset.}}
    \label{tab:baseline_eg}
\end{table*}

%%%%%%%%%%%%%%%%%%%%%%%%%%%%%%%%%%%%%%%%
% Table RU (EGTEA GAZE+)
%%%%%%%%%%%%%%%%%%%%%%%%%%%%%%%%%%%%%%%%

%\begin{table*}[!t]
%    \centering
%    \begin{adjustbox}{width=\linewidth,center}
%    \setlength{\tabcolsep}{11pt}
%    \begin{tabular}{ l c c c c c c c c} 
%        \hline
%        \multicolumn{9}{ c }{\rev{Top-5 Action Accuracy \% @ different anticipation times [$s$]}}\\
%        \hline
%        & \rev{2} & \rev{1.75} & \rev{1.5} & \rev{1.25} & \rev{1} & \rev{0.75} & \rev{0.5} & \rev{0.25} \\
%        \hline
%        \rev{RU-LSTM} & $\rev{56.82}$ & $\rev{59.13}$ & $\rev{61.42}$ & $\rev{63.53}$ & $\rev{66.40}$ & $\rev{68.41}$ & $\rev{71.84}$ & $\rev{74.28}$ \\
%        \rev{\textbf{RU-LSTM Smooth GL}} & $\mathbf{\rev{59.99}}$ & $\mathbf{\rev{62.02}}$ & $\mathbf{\rev{63.95}}$ & $\mathbf{\rev{66.47}}$ & $\mathbf{\rev{68.74}}$ & $\mathbf{\rev{72.16}}$ & $\mathbf{\rev{75.21}}$ & $\mathbf{\rev{78.11}}$ \\
%        \hline
%        \rev{Improv.} & \rev{+3.17} & \rev{+2.89} & \rev{+2.53} & \rev{+2.94} & \rev{+2.34} & \rev{+3.75} & \rev{+3.37} & \rev{+3.83} \\
%        \hline
%    \end{tabular}
%    \end{adjustbox}
%    \caption{\rev{Top-5 accuracy for action anticipation task at different anticipation time step on the EGTEA Gaze+ dataset.}}
%    \label{tab:ru_eg}
%\end{table*}

\subsection{Models and Baselines}
In the comparative analysis, we exploit the architecture proposed in Sec.~\ref{sec:model_architecture} employing the different label smoothing techniques defined in Sec.~\ref{sec:label_smoothing}.
In our experiments we consider models trained using one-hot vectors (One Hot), uniform smoothing (Smooth Uniform), temporal soft labels (Smooth TE), Verb-Noun soft labels (Smooth VN), GloVe based soft labels (Smooth GL) and GloVe + Verb-Noun soft labels (Smooth GL+VN). We design the last method (Smooth GL+VN) by smoothing hard labels with the average of the two related priors. 
%Besides our model, 
We also evaluate the effect of the proposed label smoothing techniques on the state-of-the-art approach RU-LSTM~\cite{furnari2020rulstm}.
\rev{In this case, we choose \textit{GL+VN} as main label smoothing technique.}

To implement our architecture, we used TensorFlow 2. Each model is trained for 100 epochs with batch size of 256 using Adam optimizer with learning rate of $0.001$.
The best model is selected using early stopping by monitoring the Top-5 accuracy at anticipation time $\tau_a = 1$ on the validation set. 
\medskip

\textbf{\textit{Results.}}
For each label smoothing method we select the best smooth factor $\alpha^*$ with a grid search procedure between $0$ and $1$ and step size $\Delta\alpha = 0.05$. The smooth factors used for the results discussed in this section are shown in Table~\ref{tab:alpha}.

We notice that our label smoothing procedures such as Verb-Noun, GloVe and Temporal perform well with higher smooth factors ($\alpha^*\simeq0.5$) compared to Uniform label smoothing ($\alpha^* = 0.1$). This suggests that, when soft labels encodes semantic information, prior information becomes more relevant assuming the same importance of the ground-truth due to the similar smooth factors ($\alpha \simeq 1 - \alpha$). During training, we fix $\alpha = \alpha^*$ for each method and, in order to have a robust performance estimation among trials, we iterate ten times all the experiments. 

Table~\ref{tab:baseline} reports our results on the EPIC-KITCHENS validation set. We notice that all our proposed label smoothing methods improve model performance as compared with training using one-hot encoded labels. Soft labels based on GloVe + Verb-Noun attain best performance improving the Top-5 accuracy from $+1.17\%$ to $+2.13\%$ compared to hard labels. 
To validate our results, we trained \rev{RU-LSTM} \cite{furnari2020rulstm}, the state-of-the-art model for action anticipation, considering label smoothing  using EPIC-KITCHENS validation and test sets.
\rev{Similar results are obtained with the RU-LSTM architecture using both one-hot encoding (our baseline) and smoothed labels.}
We select \textit{GloVe + Verb-Noun} soft labels since they show a higher performance increase. As shown by the experimental results \rev{(see Table~\ref{tab:baseline})}, label smoothing improves the Top-5 accuracy for all the anticipation times of a margin from $+\%0.58$ to $+1.59\%$. Such behavior highlights the systematic effect of smoothed labels on the model performance. \rev{In Table~\ref{tab:branches} we also report the performance of our label smoothing procedures for each model branch. The results show a systematic increase of the accuracy using soft labels for each branch mainly for the GloVe-based method.}

\rev{Table~\ref{tab:baseline_eg} reports our results on the EGTEA Gaze+ test set. All our label smoothing methods improve the baseline trained with one-hot labels. GloVe softening procedure shows the highest accuracy increase ranging from $+1.31\%$ to $+1.89\%$.
We also report the performance of RU-LSTM baseline trained using GloVe label smoothing and, in this case, for all the anticipation times, the accuracy is improved by a significant margin, from $+2.34\%$ to $+3.83$. Compared to the EPIC-KITCHENS dataset, the larger improvement could be owed to the smaller dataset  showing that label smoothing helps bridge the lack of data in the training set.}

In Table~\ref{tab:anticipation_ek_test}, we report the results obtained considering the test set of EPIC-KITCHENS. Different approaches are considered for the comparison.
It is wort noting that the method which consider label smoothing together with RU-LSTM improves the performances obtaining best Top-1 and Top-5 accuracy for anticipating \textit{verb}, \textit{noun} and \textit{action}. Label smoothing helps also to improve Precision for \textit{verb} and obtains comparable results in anticipating \textit{action} and \textit{noun}. Results in terms of Recall point out that label smoothing helps for \textit{noun} anticipation maintaining comparable results for the anticipation of \textit{verb} and \textit{action}.
%We show that smoothing methods improve models performance, considering both Top-5 accuracy and Top-1 accuracy reaching best results in terms of Precision and Recall compared to RU-LSM.

\begin{figure*}[ht]
    \centering
    \begin{tabular}{c c}
         \includegraphics[width=0.49\textwidth]{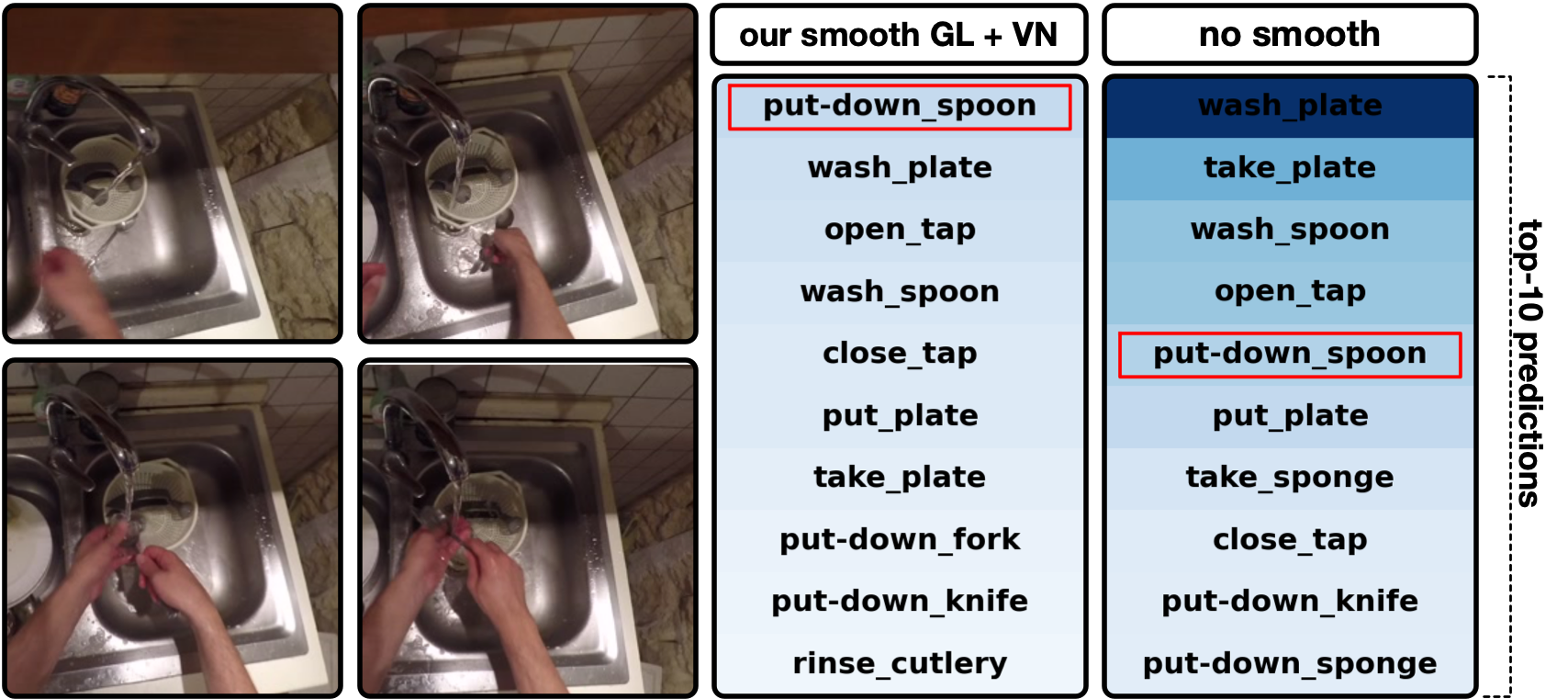} & \includegraphics[width=0.49\textwidth]{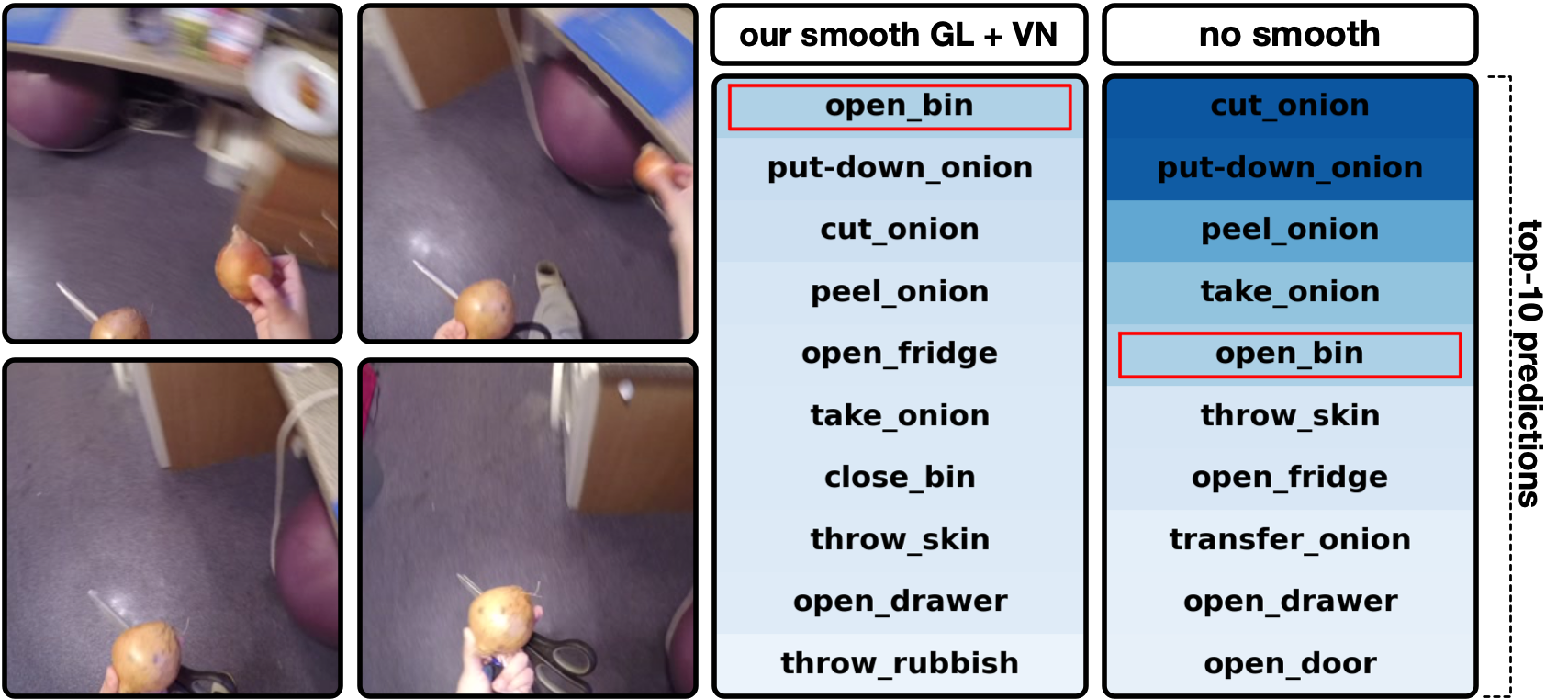}\\
        \includegraphics[width=0.49\textwidth]{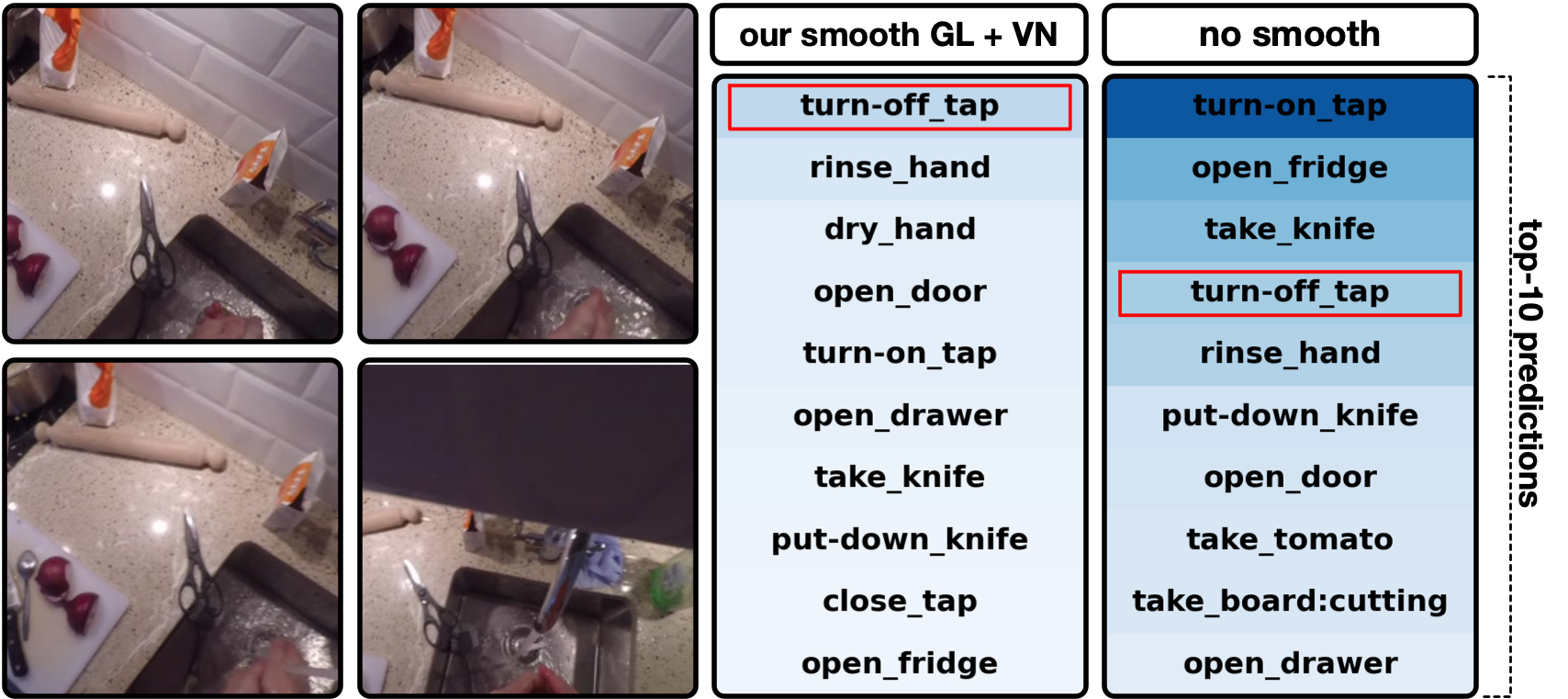} & \includegraphics[width=0.49\textwidth]{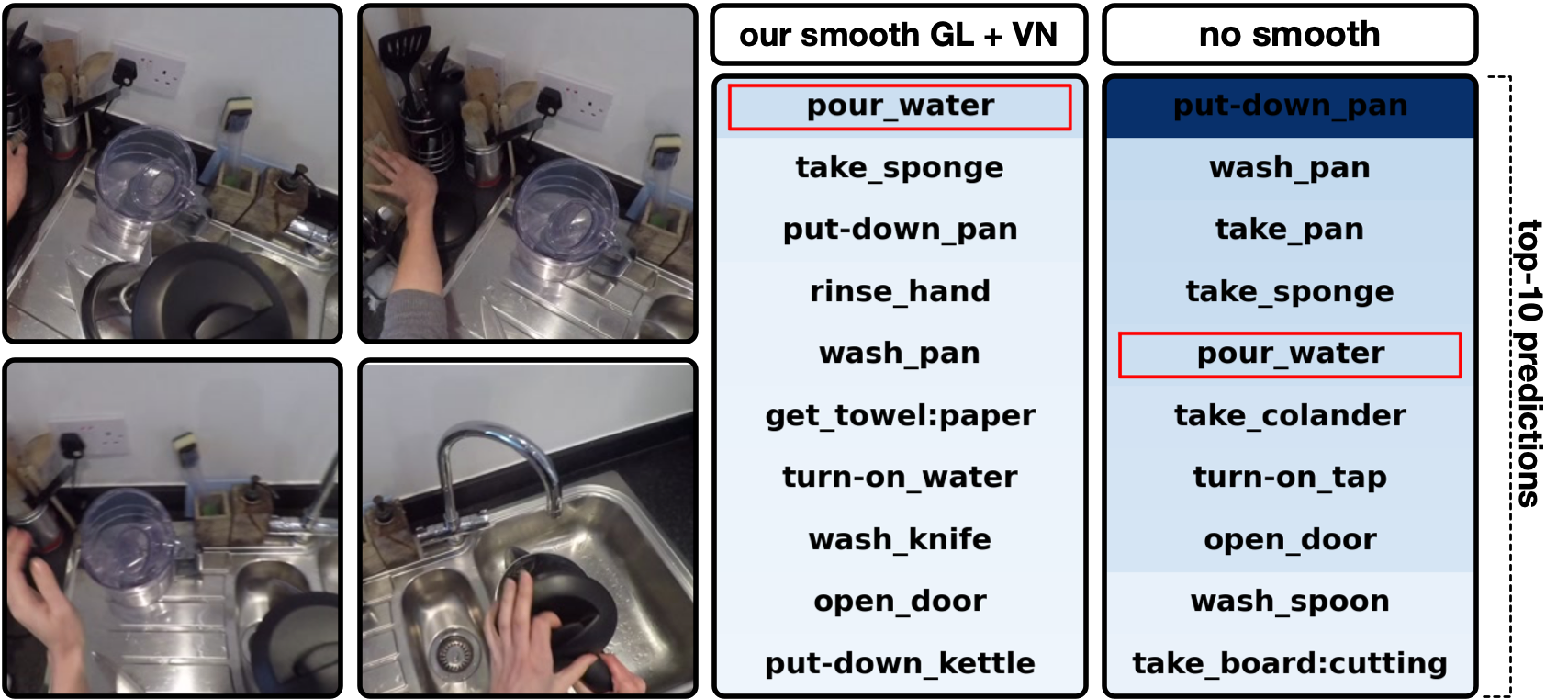}
    \end{tabular}
\caption{Qualitative results of our smoothing labels procedure. We show the comparison between the top-10 predictions of our model at $\tau_a = 1$ second trained either with smoothed labels or with one-hot vectors. These examples are from the validation set where both the models predict correctly, under the top-5 accuracy, the upcoming action. The model trained on one-hot labels is more confident and tries to concentrates all the prediction energy on a very restricted set of actions, without capturing the uncertainty of the future. By contrast, smoothing labels shape the prediction distribution considering similar actions to the ground truth.}
\label{fig:qual_res}
\end{figure*}

Finally, Fig.~\ref{fig:qual_res} shows some qualitative results obtained with our framework.%, whereas Fig.~\ref{fig:smooth_matrices} depicts prior component representations of the proposed label smoothing procedures.

%%%%%%%%%%%%%%%%%%%%%%%%%%%%%%%%%%%%%%%%
% Qualitative results
%%%%%%%%%%%%%%%%%%%%%%%%%%%%%%%%%%%%%%%%

%%%%%%%%%%%%%%%%%%%%%%%%%%%%%%%%%%%%%%%%
% Smooth matrices
%%%%%%%%%%%%%%%%%%%%%%%%%%%%%%%%%%%%%%%%

% \begin{figure*}[ht]
%     \centering
%     \begin{subfigure}[h]{0.25\textwidth}
%         \centering
%         \includegraphics[width=0.95\textwidth]{./plots/glove-sim_soft-matrix.png}
%         \caption{GloVe Smoothing Matrix.}
%         \label{fig:smooth_matrices_gl}
%     \end{subfigure}
%     \begin{subfigure}[h]{0.25\textwidth}
%         \centering
%         \includegraphics[width=0.95\textwidth]{./plots/verb-noun_soft-matrix.png}
%         \caption{Verb-Noun Smoothing Matrix.}
%         \label{fig:smooth_matrices_vn}
%     \end{subfigure}
%     \begin{subfigure}[h]{0.25\textwidth}
%         \centering
%         \includegraphics[width=0.95\textwidth]{./plots/action_adj_matrix.png}
%         \caption{Temporal Smoothing Matrix.}
%         \label{fig:smooth_matrices_te}
%     \end{subfigure}
%     \caption{\rev{Credo che con i nuovi esperimenti su GAZE+ possiamo anche togliere questa figura.}Smooth Labels Matrices. Each matrix represents the prior component for the label smoothing procedure reported in Eq.~(\ref{eq:smooth_label_gen}) and each row corresponds to a single label prior. In the GloVe and Verb-Noun priors can be recognized squared structures on the diagonal because the labels are alphabetically ordered. The Temporal prior has sparse row entries since there is no major occurrence trend or structure in the training set.}
%     \label{fig:smooth_matrices}
% \end{figure*}

\begin{table*}[t]
		\begin{adjustbox}{width=\linewidth,center}
		\setlength{\tabcolsep}{3pt}
		\begin{tabular}{llccc|ccc|ccc|ccc}
			%		\cline{2-18}
			& & \multicolumn{3}{c|}{Top-1 Accuracy\%} & \multicolumn{3}{c|}{Top-5 Accuracy\%} & \multicolumn{3}{c|}{Avg Class Precision\%} & \multicolumn{3}{c}{Avg Class Recall\%} \\ \hline
			& & VERB & NOUN & ACTION & VERB & NOUN & ACTION & VERB & NOUN & ACTION & VERB & NOUN & ACTION \\ \hline
			\multirow{7}{*}{\rotatebox{90}{{S1}}} &
			{DMR~\cite{vondrick2016anticipating}} & 26.53 & 10.43 & 01.27 & 73.30 & 28.86 & 07.17 & 06.13 & 04.67 & 00.33 & 05.22 & 05.59 & 00.47\\
			&
			{2SCNN~\cite{Damen2020Collection}} & 29.76 & 15.15 & 04.32 & 76.03 & 38.56 & 15.21 & 13.76 & 17.19 & 02.48 & 07.32 & 10.72 & 01.81\\
			&ATSN~\cite{Damen2020Collection} & {31.81} & 16.22 & 06.00 & {76.56} & 42.15 & {28.21} & {23.91} & {19.13} & 03.13 & 09.33 & 11.93 & 02.39\\
			&MCE~\cite{Furnari_2018_ECCV_Workshops} & 27.92 & 16.09 & {10.76} & 73.59 & 39.32 & 25.28 & 23.43 & 17.53 & {06.05} & {14.79} & 11.65 & {05.11}\\
			&{ED~\cite{gao2017red}} & 29.35 & 16.07 & 08.08 & 74.49 & 38.83 & 18.19 & 18.08 & 16.37 & 05.69 & 13.58 & {14.62} & 04.33\\
			&{Miech et al.~\cite{miech2019leveraging}} & 30.74 & {16.47} & 09.74 & 76.21 & {42.72} & 25.44 & 12.42 & 16.67 & 03.67 & 08.80 & 12.66 & 03.85\\
			&RU-LSTM~\cite{furnari2020rulstm} &
			{33.04} & {22.78} & {14.39} & {79.55} & {50.95} & {33.73} & {25.50} & {24.12} & \textbf{07.37} & \textbf{15.73} & {19.81} & \textbf{07.66}\\
			\hline
			&\textbf{RU-LSTM GL+VN Smoothing} & \textbf{35.04} & \textbf{23.03} & \textbf{14.43} &	\textbf{79.56} & \textbf{52.90} & \textbf{34.99} & \textbf{30.65} &	\textbf{24.29} & {06.64} & 14.85 & \textbf{20.91} &	07.61\\ \hline
			\hline
			\multirow{7}{*}{\rotatebox{90}{{S2}}} &
			DMR~\cite{vondrick2016anticipating} & 24.79 & 08.12 & 00.55 & 64.76 & 20.19 & 04.39 & 09.18 & 01.15 & 00.55 & 05.39 & 04.03 & 00.20\\
			&{2SCNN~\cite{Damen2020Collection}} & 25.23 & 09.97 & 02.29 & {68.66} & 27.38 & 09.35 & {16.37} & 06.98 & 00.85 & 05.80 & 06.37 & 01.14\\
			&ATSN~\cite{Damen2020Collection} & {25.30} & {10.41} & 02.39 & 68.32 & {29.50} & 06.63 & 07.63 & {08.79} & 00.80 & 06.06 & {06.74} & 01.07\\
			&MCE~\cite{Furnari_2018_ECCV_Workshops} & 21.27 & 09.90 & {05.57} & 63.33 & 25.50 & {15.71} & 10.02 & 06.88 & {01.99} & {07.68} & 06.61 & {02.39}\\
			&{ED~\cite{gao2017red}} & 22.52 & 07.81 & 02.65 & 62.65 & 21.42 & 07.57 & 07.91 & 05.77 & 01.35 & 06.67 & 05.63 & 01.38\\
			&{Miech et al.~\cite{miech2019leveraging}} & {28.37} & {12.43} & {07.24} & {69.96} & {32.20} & {19.29} & 11.62 & 08.36 & {02.20} & {07.80} & {09.94} & {03.36}\\
			&RU-LSTM~\cite{furnari2020rulstm} & {27.01} & {15.19} & {08.16} & {69.55} & {34.38} & {21.10} & {{13.69}} & \textbf{09.87} & {03.64} & \textbf{09.21} & {11.97} & \textbf{04.83}\\
			\hline
			&\textbf{RU-LSTM GL+VN Smoothing} & \textbf{29.29} & \textbf{15.33} & \textbf{08.81} & \textbf{70.71} &	\textbf{36.63} &	\textbf{21.34} &	\textbf{14.66} &	09.86 &	\textbf{04.48} &	08.95 &	\textbf{12.36} &	04.78 \\ \hline
		\end{tabular}
	\end{adjustbox}	
	\caption{Results of action anticipation on the test sets of EPIC-Kitchens. The test set is divided into kitchens already seen $S1$ or unseen $S2$ from the model. The table confirms also on the test set that our smoothing labels procedure can improve performances of the state-of-the-art model RU-LSTM.}
	\label{tab:anticipation_ek_test}
\end{table*}

%------------------------------------------------------------------
\section{Conclusion}
\label{sec:conclusion}
\rev{This study proposed a knowledge distillation procedure
which accounts for multi-modality of future predictions in the action anticipation task. We implemented knowledge distillation through label smoothing by relying on priors capturing inter-dependencies between verb and noun labels, past and future actions, as well as semantic relevance between different actions.}
%This study proposed a knowledge distillation procedure via label smoothing for leveraging the multi-modal future component of the action anticipation problem. We generalized the idea of label smoothing by designing semantic priors of actions that are used during training as ground truth labels. We implemented a LSTM baseline model that can anticipate actions at multiple time steps starting from multi-modal representation of the input video. 
Experimental results corroborate out findings compared to state-of-the-art models highlighting that label smoothing systematically improves performance when dealing with future uncertainty.

\medskip
\paragraph*{Acknowledgements}
Research at the University of Padova is partially supported by MIUR PRIN-2017 PREVUE grant. Authors from the Univ. of Padova gratefully acknowledge the support of NVIDIA for their donation of GPUs, and the UNIPD CAPRI Consortium for its support and access to computing resources.
Research at the University of Catania is supported by PTR 2016-2018 Linea 2 of DMI and MIUR AIM - Linea~1 - AIM1893589 - CUP E64118002540007.


%\section*{Acknowledgment}





\balance
\bibliographystyle{IEEEtran}
\bibliography{IEEEabrv,actions}

\end{document}